\documentclass{article}

\usepackage{microtype}
\usepackage{graphicx}
\usepackage{subcaption}
\usepackage{booktabs} 

\usepackage{hyperref}

\usepackage[accepted]{icml2023}

\usepackage{amsmath}
\usepackage{amssymb}
\usepackage{mathtools}
\usepackage{amsthm}
\usepackage{float}
\usepackage{cuted}
\usepackage{caption}
\usepackage{subcaption}
\usepackage{xcolor}
\usepackage[capitalize,noabbrev]{cleveref}

\theoremstyle{plain}

\theoremstyle{definition}

\theoremstyle{remark}

\usepackage[textsize=tiny]{todonotes}

\newcommand\newcite[1]{\citeauthor{#1} (\citeyear{#1})}

\icmltitlerunning{Concept-aware clustering for decentralized deep learning under temporal shift}

\begin{document}

\twocolumn[
\icmltitle{Concept-aware clustering for decentralized deep learning under temporal shift}



\icmlsetsymbol{equal}{*}

\begin{icmlauthorlist}
\icmlauthor{Marcus Toftås}{equal,chalmers}
\icmlauthor{Emilie Klefbom}{equal,chalmers}
\icmlauthor{Edvin Listo Zec}{equal,rise,kth}
\icmlauthor{Martin Willbo}{rise}
\icmlauthor{Olof Mogren}{rise}

\end{icmlauthorlist}

\icmlaffiliation{chalmers}{Chalmers University of Technology}
\icmlaffiliation{rise}{RISE Research Institutes of Sweden}
\icmlaffiliation{kth}{KTH Royal Institute of Technology}

\icmlcorrespondingauthor{Edvin Listo Zec}{edvin.listo.zec@ri.se}

\icmlkeywords{Decentralized learning, ICML}

\vskip 0.3in
]

\iftrue
\begin{figure*}[t!]
    \centering
    \includegraphics[width=1\textwidth]{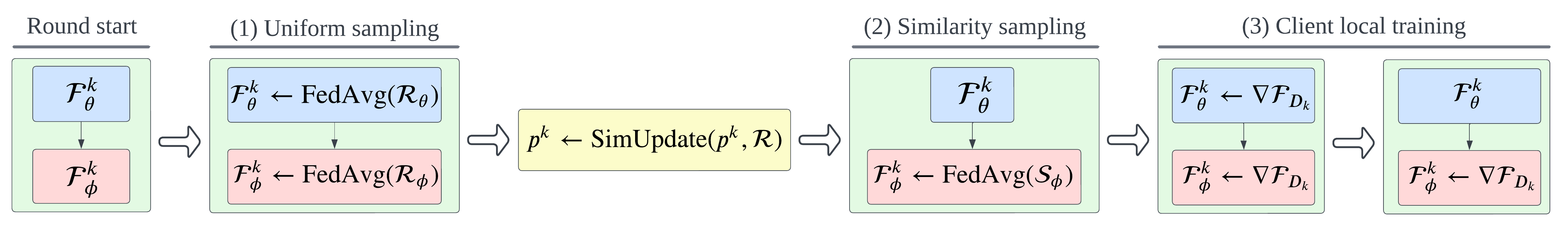}
    \caption{An illustration of the proposed solution. Each client $k$ has a model consisting of a \textcolor{blue}{feature extractor $\mathcal{F}^k_\theta$} and a \textcolor{red}{classifier $\mathcal{F}^k_\phi$}. The training consists of three steps: (1) Uniform sampling: A subset of clients $\mathcal{R}$ is randomly selected and client $k$ updates both model components using \textsc{FedAvg} with $\mathcal{R}$. (2) Similarity-based sampling: A subset of clients $\mathcal{S}$ is selected based on their similarity $s$ to client $k$, which updates only the classifier component using \textsc{FedAvg} with $\mathcal{S}$. (3) Local training: Client $k$ trains both model components locally and fine-tunes the classifier component.}
    \label{fig:alg_outline}
\end{figure*}
\fi
%




\printAffiliationsAndNotice{\icmlEqualContribution} 

\begin{abstract}
Decentralized deep learning requires dealing with non-iid data across clients, which may also change over time due to temporal shifts. While non-iid data has been extensively studied in distributed settings, temporal shifts have received no attention. To the best of our knowledge, we are first with tackling the novel and challenging problem of decentralized learning with non-iid and dynamic data. We propose a novel algorithm that can automatically discover and adapt to the evolving concepts in the network, without any prior knowledge or estimation of the number of concepts. We evaluate our algorithm on standard benchmark datasets and demonstrate that it outperforms previous methods for decentralized learning.
\end{abstract}

\section{Introduction}
The proliferation of smartphones and other devices that can continuously collect and transmit data has substantially increased the amount of data available for machine learning-based applications. However, sharing data explicitly may not be possible or desirable in some scenarios. For instance, users and businesses may have privacy concerns preventing disclosure of private or sensitive data. Alternatively, legal regulations such as the GDPR \cite{voigt2017eu} and other data protection acts may prohibit data sharing. Moreover, practical limitations such as large distributed datasets or low network bandwidth may make transmitting data to a centralized location infeasible. 
In such scenarios, distributed frameworks like federated learning (FL) can be a viable solution, as they communicate model parameters instead of data, and where the models are aggregated using federated averaging (\textsc{FedAvg}) \cite{mcmahan2017communication}. FL has already demonstrated its scalability and applicability in various domains, such as hospitals \cite{dayan2021federated}, retail stores \cite{yang2019federated}, and at companies such as Google \cite{mcmahan2022federated}.

In the FL framework multiple clients collaborate to train a shared model without exchanging their local data. However, FL often relies on a central node to coordinate the communication among the clients, which can cause communication bottlenecks and single points of failure. To overcome these limitations, decentralized learning proposes a peer-to-peer communication protocol that eliminates the need for a central node and reduces the vulnerability and computational load of any single node. However, decentralized learning also poses new challenges, such as how to optimize the client models in a decentralized manner.

Traditionally, decentralized machine learning has used consensus optimization, where clients aim to agree on a common model using a gossip learning approach \cite{kempe2003gossip,boydgossip,blotgossip}. This works well when the data distributions across clients are similar or identical. However, when the data distributions or tasks differ significantly across clients, consensus optimization can be detrimental. Therefore, recent work has suggested viewing decentralized learning as a clustering problem, where clients try to find suitable collaborators in a network of peers and avoid merging their models with dissimilar ones \cite{onoszko2021decentralized,li2022mining,listozecdecentralized}. All of these works consider non-iid data in decentralized deep learning, however they still assume that the data distributions are stationary in time. For real world scenarios, this is often not the case for edge devices that continuously collect new data. 

This work presents the first study of decentralized deep learning with temporal shifts, which account for the dynamic nature of data distributions over time. Our main contribution is an algorithm that allows clients to learn personalized models in non-iid settings where their concepts may evolve over time. Our problem setting is related to the recent work of \cite{jothimurugesan2023federated}, which investigates non-iid data across clients and time in federated learning with a central server.

\section{Background}
\subsection{Problem formulation}
Decentralized learning involves clients solving their own optimization problems, such as supervised learning. Previous work has shown that clients can benefit from communicating models with other clients who have similar tasks \cite{listozecdecentralized}. However, the authors assumed that the data distributions of each client are stationary. We address the challenges that emerge when the data distributions of clients vary over time. Our main contribution is to demonstrate that existing decentralized learning methods are not resilient to temporal shifts, and to propose a simple solution based on novel hierarchical model aggregation. We empirically show that our solution leads to improved performance for this problem.

We consider an empirical risk minimization (ERM) setup with $K$ clients that communicate (synchronously) in a peer-to-peer network, where any pair of clients can communicate at each communication round. Each client $k$ has a (private) training set $\mathcal{D}^t_k(x,y)$ generated by the underlying distribution $p^t_k(x,y)$ at time step $t$. We assume that the data is non-iid both across clients and over time, which is more realistic than the common assumption of only having non-iid data across clients. We follow \newcite{jothimurugesan2023federated} and \newcite{gama2014survey} and say that there is a concept shift at client $k$ if $p_k^t \neq p_k^{t-1}$.

This work aims to design an algorithm that can handle shifts in a distributed setting. Concept shift can be caused by various types of shifts over time. In this paper, we focus on two main types of shifts: \textit{covariate shift} (where the input distribution $p(x)$ changes but the label distribution $p(y|x)$ is stationary) and \textit{label shift} (where the label distribution $p(y)$ changes but the distribution $p(x|y)$ is stationary.) \cite{kairouz2021advances}.

\subsection{Motivation}

Decentralized learning is appealing in situations where a central server is undesirable; this could be due to privacy concerns (lack of trust in the central server) or scalability issues (central server being a bottleneck). For example, consider $K$ users with their own data distributions. A possible scenario is $K$ smartphones that collect images that reflect the users’ preferences and activities. Rather than relying on a central server, each user can communicate with peers in a network to jointly solve some optimization problem, such as learning a supervised image recognition task. However, the data distributions may differ across clients due to their personal preferences. In realistic scenarios, the images collected may also vary over time, which has not been studied in decentralized peer-to-peer learning before. Therefore, the goal of this work is to study distributed temporal shifts.

\section{Method}
We model each client as having a \textit{concept} that reflects its true optimization objective at any point in time. A client benefits from communicating with other clients that have the same concept as itself (i.e. similar data distribution), since \textsc{FedAvg} becomes detrimental otherwise \cite{mcmahan2017communication}. This can thus be framed as a clustering problem, where each client tries to find clients in the network which have a similar concept. However, the clustering is challenging because the clients cannot share their local data due to privacy concerns, and can only exchange gradients (or model parameters).

\subsection{Hierarchical aggregation with similarity based tuning (HAST)}
We propose a novel clustering algorithm that can adapt to different concepts that emerge at different times. Our algorithm, called \textbf{Hierarchical Aggregation with Similarity based Tuning} (HAST), extends the Decentralized Adaptive Clustering (DAC) method \cite{listozecdecentralized} by incorporating a hierarchical structure and a tuning mechanism based on empirical training loss. We demonstrate the effectiveness of our algorithm on two scenarios: one with two concepts and one with four concepts. We show that our algorithm outperforms DAC and other baselines in terms of accuracy and robustness under concept shift.

HAST works as follows. Each client $k$ has a neural network consisting of a feature extractor $\mathcal{F}^k_\theta$ and a classifier $\mathcal{F}^k_\phi$ (the whole model being $\mathcal{F}^k_\phi \circ \mathcal{F}^k_\theta$). The training procedure in HAST consists of three steps, illustrated in figure \ref{fig:alg_outline}. For each client $k$ and for each communication round:
\begin{enumerate}
    \item A subset of clients $\mathcal{R}$ is sampled uniformly and at random. Client $k$ updates all layers $\mathcal{F}^k_\phi \circ \mathcal{F}^k_\theta$ as in \textsc{FedAvg}.
    \item A subset of clients $\mathcal{S}$ is randomly sampled based on their similarity to client $k$. Client $k$ updates \textit{only} the classifier layers $\mathcal{F}^k_\phi$ using average aggregation.
    \item Client $k$ performs local training on its own data, updating the whole model $\mathcal{F}^k_\phi \circ \mathcal{F}^k_\theta$. Afterwards, it fine-tunes only the classifier $\mathcal{F}^k_\phi$.
\end{enumerate}
The similarity function is the same as used by \cite{listozecdecentralized}, based on the empirical training loss of client model $i$ on the data of client $j$: $s_{ij} = 1/\ell(w_i;x_j)$. This similarity score is transformed using a softmax with a temperature scaling $\tau$ in order to get a probability vector $\tilde{s}_{ij}$ for each client and each communication round.

\subsection{Experimental setup}
Our goal is to develop an algorithm that is robust to distributional shifts (not decreasing significantly in accuracy between tasks). To demonstrate the challenge of decentralized shift, we measure test accuracy over time and study how robust different methods are to temporal shifts on two computer vision datasets. Our code is found on Github. \footnote{\url{https://github.com/EmilieKar/HAST}}

To investigate the effects of \textit{covariate shift}, we consider the PACS dataset \cite{li2017deeper} which is typically used for domain adaptation. It consists of four domains with the same seven labels in each domain. We simulate covariate shift by changing domains for a client $k$ over time $t$, i.e. $\mathcal{D}^t_k(x)$ varies over time but keeping $\mathcal{D}^t_k(y|x)$ fixed. To investigate the effects of \textit{label shift}, we consider the CIFAR-10 dataset \cite{krizhevsky2009learning}. We simulate label shift by varying $\mathcal{D}_k^t(y)$ over time but keeping $\mathcal{D}
^t_k(x|y)$ fixed, creating two clusters based on the labels: one animal cluster (four labels) and one vehicle cluster (four labels).

\textbf{Baselines.} We compare our proposed method to two main baselines. The first we refer to as \textit{Random}, where all clients communicate randomly using a gossip protocol. The second is \textit{DAC} \cite{listozecdecentralized}, where clients communicate using the similarity metric based on empirical training loss. For a fair comparison with the baselines, we allow all methods to sample the same number of clients per communication round. Since HAST performs two stages of aggregation (random and similarity-based), it effectively samples $2n$ clients, where $n=3$ in this paper. Therefore, we allow the baselines to also sample $2n$ clients for each round.

\textbf{Hyperparameters}. We performed a grid search over the hyperparameters of each baseline and selected the ones that achieved the highest validation accuracy. We also verified that the optimal learning rate was not at the boundary of the searched grid.

\textbf{Models.} We use a small CNN model for each client in all experiments. The model has two convolutional layers ($\mathcal{F}_\theta$), two fully-connected layers and an output layer ($\mathcal{F}_\phi$). The model is not designed to achieve state-of-the-art performance on the supervised tasks, but rather to have enough capacity to solve the tasks while exploring decentralized temporal shifts. The number of clients is set to 50 in the CIFAR-10 experiments, and 20 in the PACS experiments.

\begin{figure*}[t]
     \centering
     \begin{subfigure}{0.24\textwidth}
         \centering
         \includegraphics[width=\textwidth]{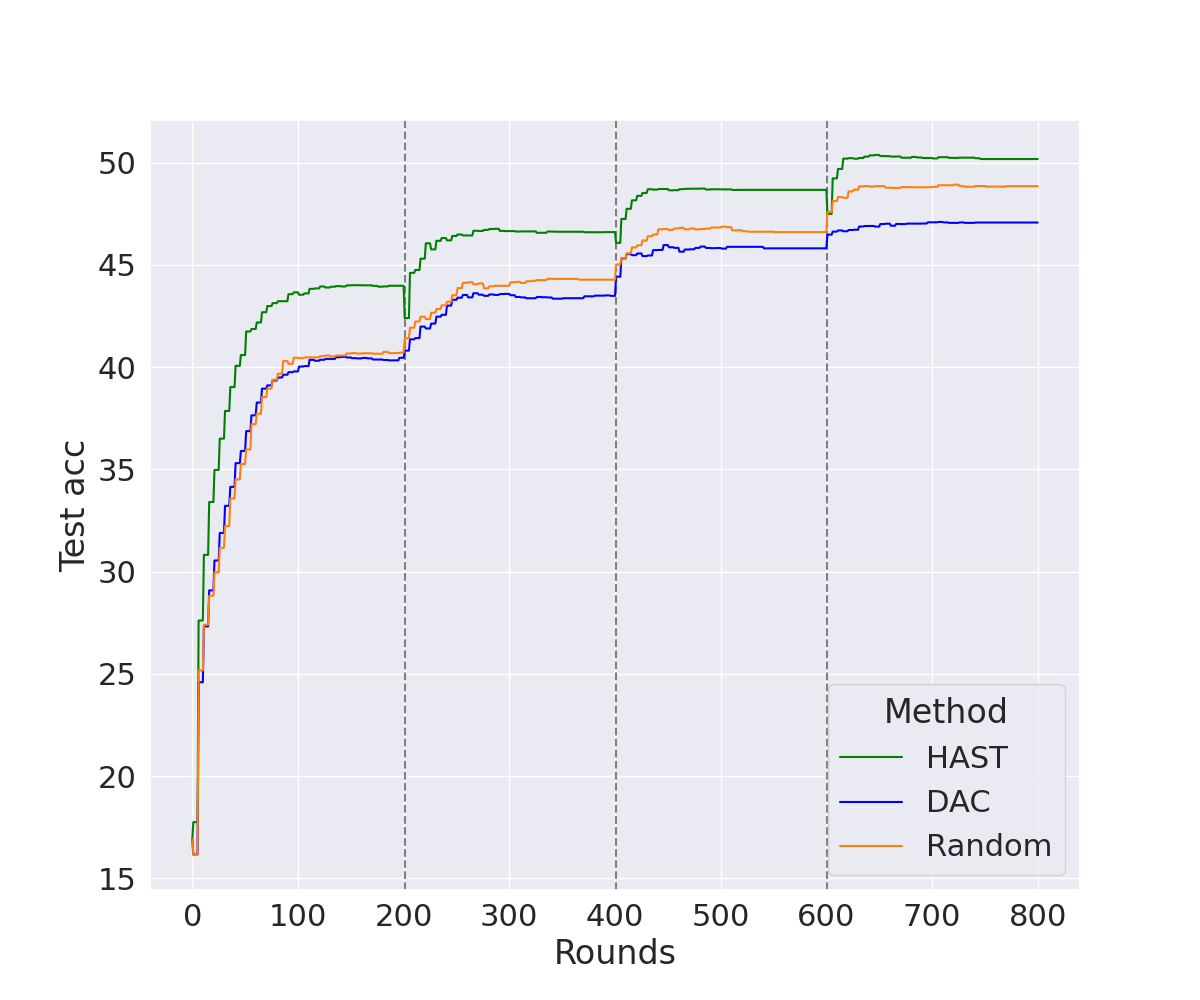}
         \caption{$C=4$.}
         \label{fig:pacs}
     \end{subfigure}
     \hfill
     \begin{subfigure}{0.24\textwidth}
         \centering
         \includegraphics[width=\textwidth]{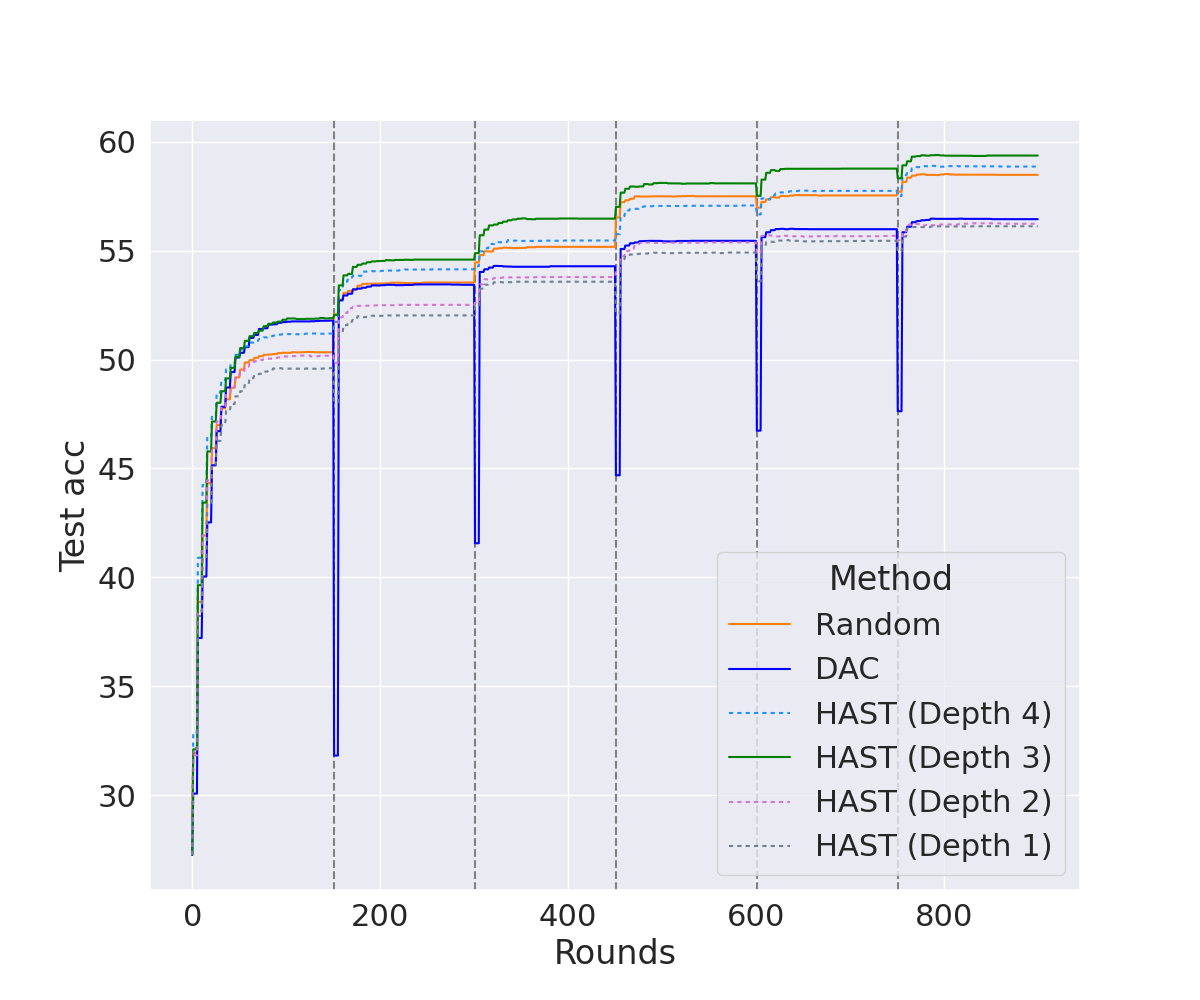}
         \caption{$C=2$, animal/vehicles.}
         \label{fig:cifar1}
     \end{subfigure}
     \hfill
     \begin{subfigure}{0.24\textwidth}
          \centering
         \includegraphics[width=\textwidth]{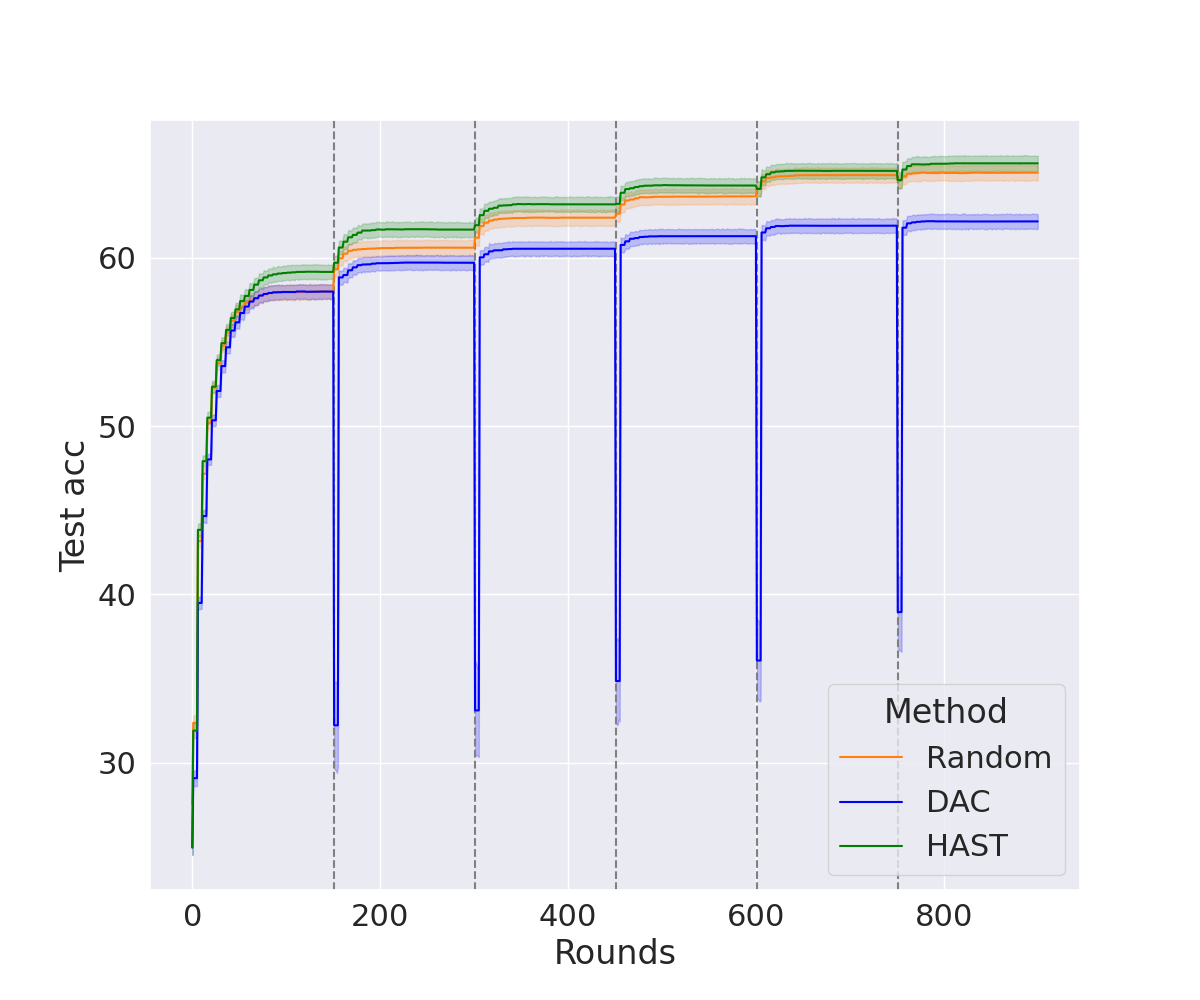}
         \caption{$C=2$, random labels.}
         \label{fig:cifar2}
     \end{subfigure}
     \hfill
     \begin{subfigure}{0.24\textwidth}
          \centering
         \includegraphics[width=\textwidth]{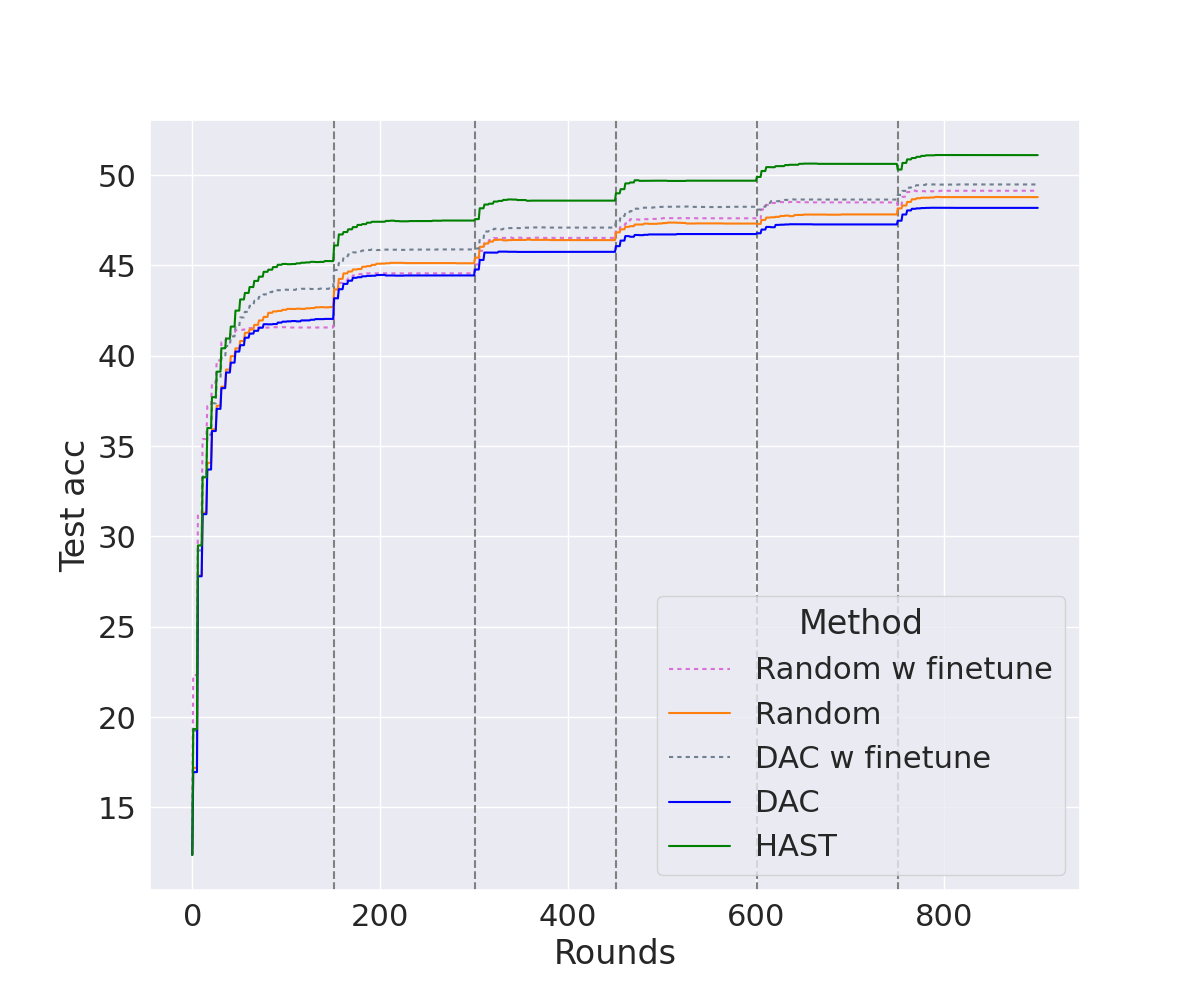}
         \caption{$C=1$.}
         \label{fig:cifar3}
     \end{subfigure}
        \caption{Test accuracy as a function of communication rounds for varying datasets and number of clusters $C$. (a): PACS, (b-d): CIFAR-10.}
        \label{fig:three graphs}
\end{figure*}

\section{Results and discussion}
Firstly, we evaluate our method on the PACS dataset, which consists of four domains: photo, art painting, cartoon, and sketch. We simulate a dynamic environment where clients may experience domain shifts over time. Specifically, at a certain communication round (marked by the vertical dashed line in figure \ref{fig:pacs}), each client randomly switches to a different domain with probability $\frac{3}{4}$. Figure \ref{fig:pacs} shows the mean test accuracy over clients as a function of communication rounds. We can see that our method is robust to domain shifts and outperforms the baselines. Interestingly, the DAC baseline performs even worse than the Random baseline. We attribute this to overfitting, as DAC merges all layers of the model with clients selected based on the similarity score. In contrast, HAST first aggregates the model with random clients and then only aggregates the classifier layer with similar clients, resulting in a more generalizable model.

We conduct experiments to investigate the effect of the aggregation scheme in HAST on the performance of personalized models. We vary the number of layers that are aggregated in the second step of HAST using the similarity sampling and compare it with DAC and Random. Figure \ref{fig:cifar1} shows the results on the CIFAR-10 dataset with $C=2$ clusters divided into animals and vehicles. The depth parameter indicates how many layers are included in the aggregation
. For example, a depth of 4 means that all layers except the first one are aggregated, while a depth of 1 means that only the output layer is aggregated. We conclude that the optimal depth for HAST is 3, which is used in all other experiments. 

Figure \ref{fig:cifar2} presents the results on CIFAR-10 with two clusters that have random labels. This setting creates a smaller distribution gap between the two clusters (compared to the setting with animals vs vehicles), which makes the Random baseline more competitive. Nevertheless, HAST still surpasses both baselines in this setting. We also note that unlike DAC, which suffers from severe overfitting to the current domain and shows a large performance drop under domain shifts, HAST maintains its robustness across domains.

We perform an ablation study to investigate the effect of the third step of HAST, where we finetune the classifier layer. We compare HAST with DAC and Random, both with and without finetuning, on the CIFAR-10 dataset with one cluster. This means that the data is iid among clients. Figure \ref{fig:cifar3} shows the results of this experiment. We observe that finetuning improves the performance of all methods and that HAST outperforms both baselines, regardless of finetuning.

\section{Conclusions}
We have presented a novel aggregation method for decentralized deep learning that can cope with temporal shifts in a peer-to-peer network. This is the first work to address this problem and to demonstrate the benefits of aggregating different layers of a neural network using \textsc{FedAvg} for robustness and generalization under temporal shifts. Our proposed algorithm uses soft clustering to group clients with similar concepts and allows them to update their beliefs of potential collaborators over time. This enables clients to smoothly transition between different collaboration groups as their concepts change. Moreover, our algorithm employs a two-stage aggregation scheme that makes the personalized models robust to concept changes by leveraging the knowledge from other clients.

\section{Related work}
Gossip learning is a peer-to-peer communication protocol that has been studied in various machine learning settings \cite{kempe2003gossip,boydgossip,ormgossip} and applied in a decentralized deep learning setup to learn personalized models \cite{blotgossip}. However, gossip learning is not suitable for situations where client data is distributed non-iid, as it assumes all clients share the same objective.

Previous work in federated learning has addressed this problem by introducing multiple central models to which clients are assigned, and clustering clients with similar objectives \cite{ghosh2020efficient}. However, this approach relies on determining the appropriate number of central models on beforehand and assigns clients to hard clusters, which may not capture the overlapping client interests.

\newcite{listozecdecentralized} proposed an algorithm that assigns clients to soft clusters. These soft cluster assignments are continuously learned over communication rounds based on client similarity, which is approximated by training losses as by \newcite{onoszko2021decentralized}. This work tackles many challenges in decentralized learning where data is non-iid distributed among clients, both in terms of cluster sizes and different types of distributional shifts. However, they do not consider time-dependent distributional shifts.

\newcite{jothimurugesan2023federated} addresses federated learning with distributed concept shift, where clients have different and dynamic data distributions. They present FedDrift, which learns multiple global models for different concepts. They are the first to tackle FL with concept shift, challenging the single-model paradigm. The work treats shift adaptation as a time-varying clustering problem, and uses hierarchical clustering to handle an unknown number of concepts. 

Similarly to \newcite{jothimurugesan2023federated}, we also view concept shift adaptation as a time-varying clustering problem. However, our solution differs in several aspects. First, we consider a decentralized learning framework, where there is no central server. Second, we build on previous work on round-based client similarity approximation and soft-cluster assignment, enabling clients to aggregate their models with similar peers. Third, we introduce a novel hierarchical model aggregation step, which ensures that every client in the network can perform well on their own current distribution and also adapt quickly to new concepts
.






\bibliography{example_paper.bib}
\bibliographystyle{icml2023}

\newpage
\appendix
\onecolumn
\section{Appendix}
\subsection{Concept shift experiment}
\begin{figure}[h!]
     \centering
     \begin{subfigure}{0.33\textwidth}
         \centering
         \includegraphics[width=\textwidth]{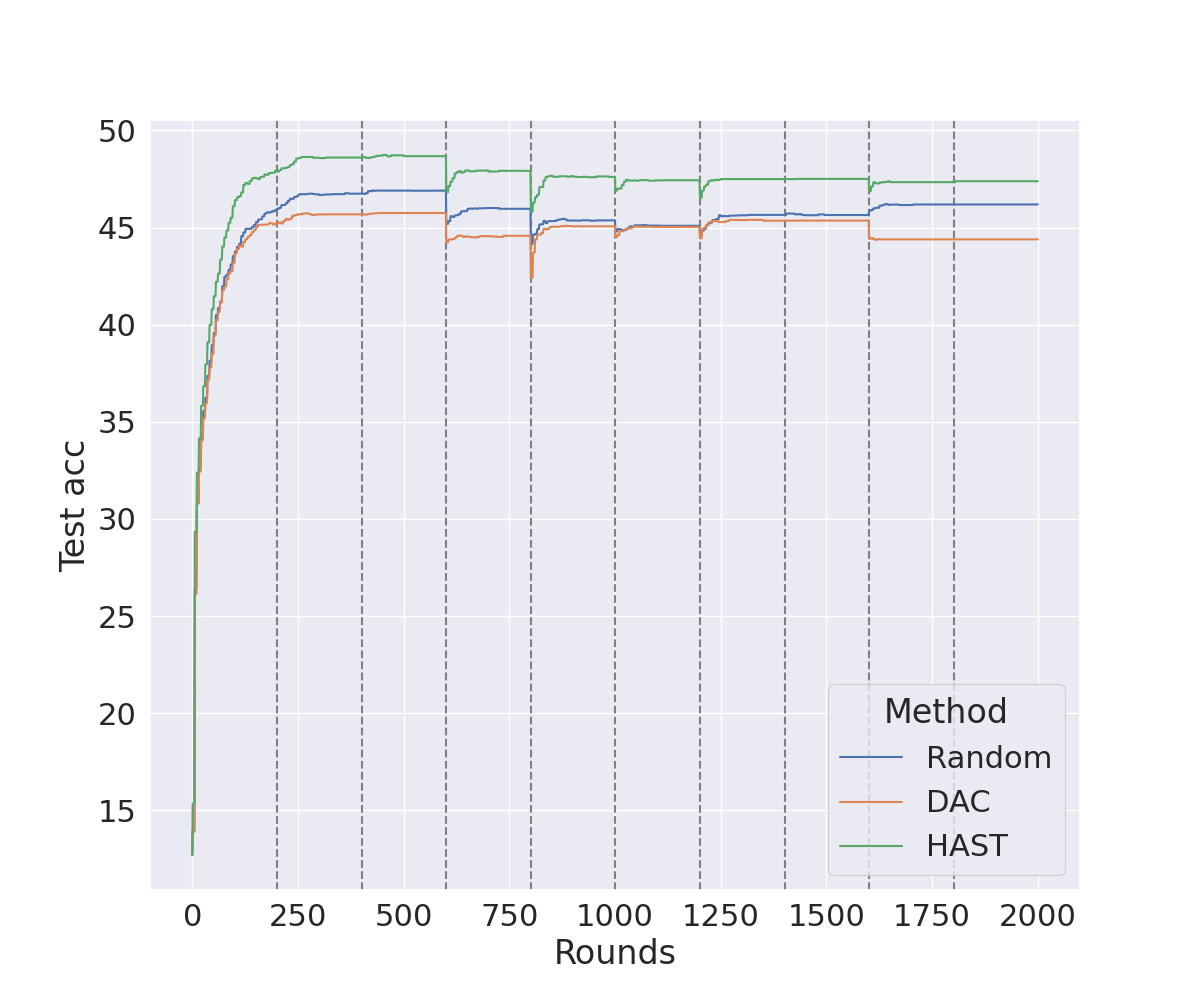}
         \caption{$C=2$, FedDrift pattern.}
         \label{fig:swap_2}
     \end{subfigure}
     \hfill
     \begin{subfigure}{0.33\textwidth}
         \centering
         \includegraphics[width=\textwidth]{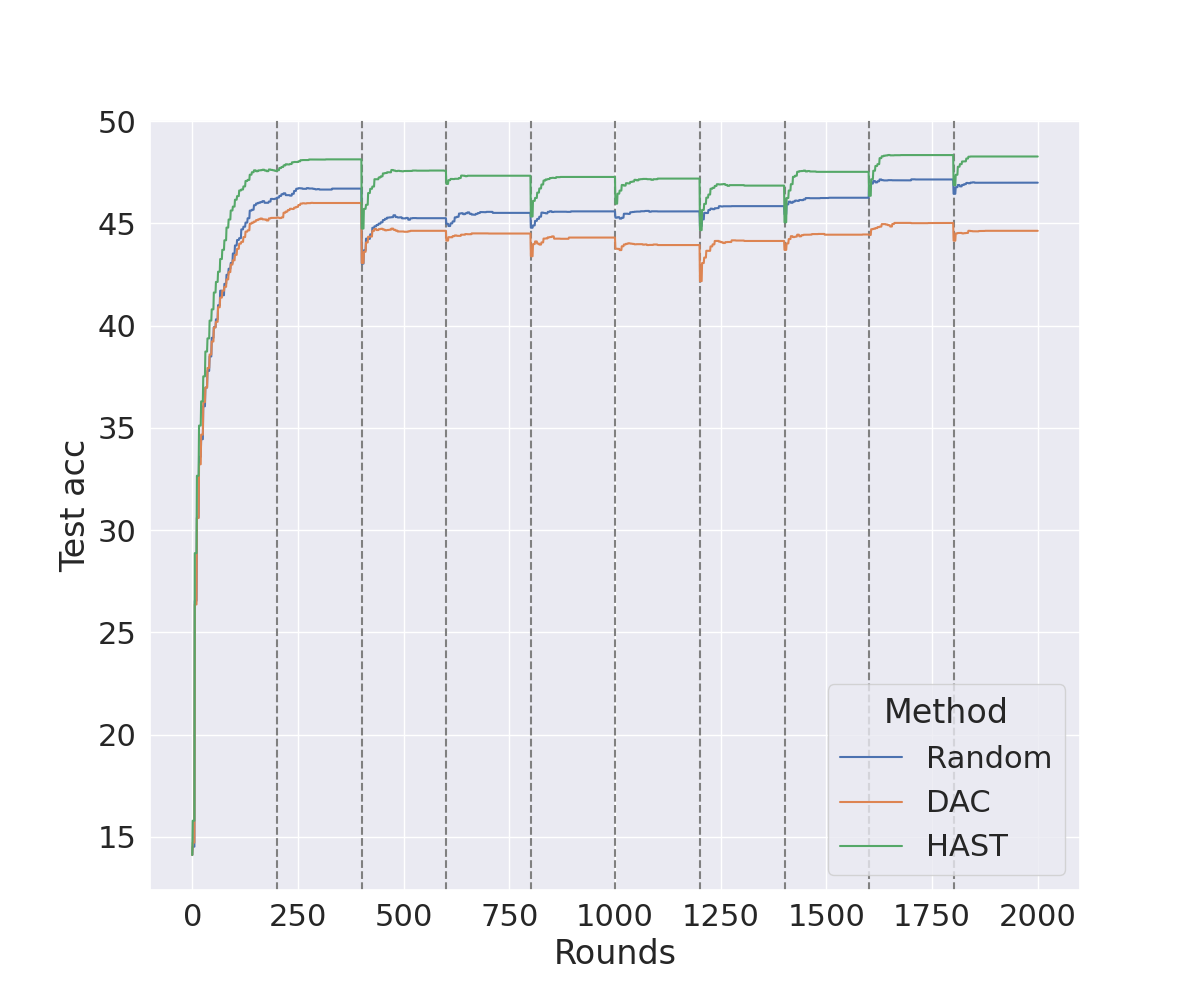}
         \caption{$C=4$, FedDrift pattern.}
         \label{fig:swap_4}
     \end{subfigure}
     \hfill
     \begin{subfigure}{0.33\textwidth}
          \centering
         \includegraphics[width=\textwidth]{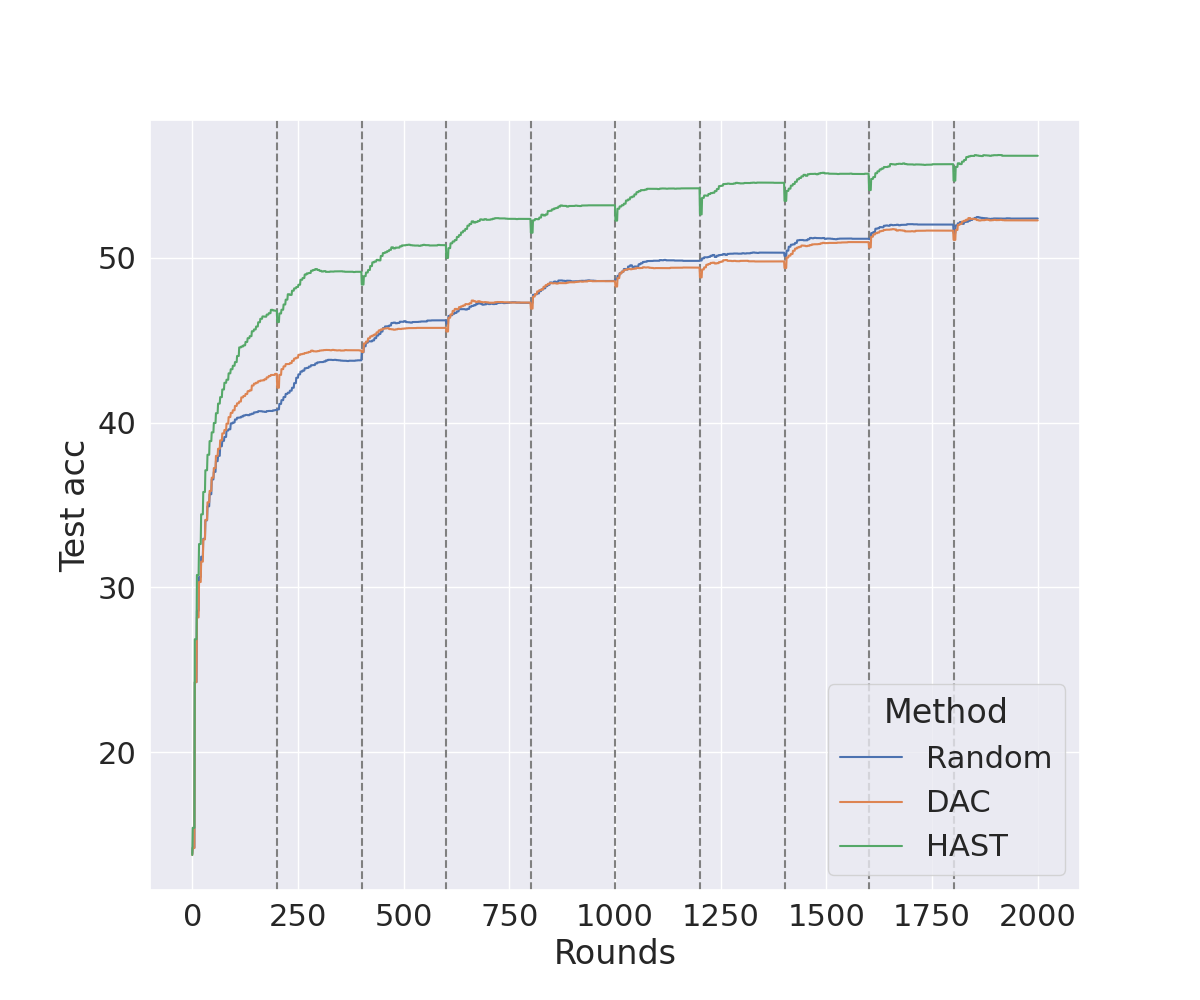}
         \caption{$C=4$, Random pattern.}
         \label{fig:swap_4_rand}
     \end{subfigure}
     \hfill
     \caption{Test accuracy as a function of communication rounds for varying number of clusters and client concept shift patterns. CIFAR-10.}
     \label{fig:swap}
 \end{figure}

In addition to the experiments shown in the main paper we also evaluated our algorithm on the concept shift \textit{label swap}, that is shifts where $p^t_k(y|x) \neq p^{t-1}_k(y|x)$. Not to be confused with the more general term we defined in the paper meaning simply that $p^t_k \neq p^{t-1}_k$.

\textbf{Experimental setup.} We grouped the clients into either two or four clusters. In Figures \ref{fig:swap_2} and \ref{fig:swap_4}, we used 20 clients in total, while in Figure \ref{fig:swap_4_rand}, we used 50 clients. The label distribution across the clusters was consistent in all experiments, with eight distinct labels. The label distribution across the clusters remained consistent in all experiments, consisting of eight distinct labels. However, in the two-cluster experiment, a \textit{label swap} was simulated by swapping two labels within one of the clusters. Similarly, in the four-cluster experiments, the label swaps within clusters were conducted with care to ensure non-overlapping exchanges of labels between the clusters. Note that Figures \ref{fig:swap_2} and \ref{fig:swap_4} followed a fixed client shift pattern as described in previous work \cite{jothimurugesan2023federated}, specifically Figures 2 and 3 in their paper. On the other hand, Figure \ref{fig:swap_4_rand} randomly assigned clients to clusters, similar to our main paper.

\textbf{Result.} Figure \ref{fig:swap} shows the results of the three \textit{label swap} experiments, where HAST outperforms all the baselines. In Figure \ref{fig:swap_2}, we see that HAST achieves the best performance in the two-cluster experiment, regardless of the client shift phase. HAST performs better than other methods throughout the experiment. It does well in the initial phase when all clients share the same cluster, in the middle phase when clients start to move to the second cluster, and in the final phase when all clients are in the second cluster only. Similarly, in Figure \ref{fig:swap_4}, HAST dominates the four-cluster experiment following the FedDrift pattern. Finally, in Figure \ref{fig:swap_4_rand}, HAST shows a significant advantage over the baselines in the random cluster assignment experiment.

\end{document}